\documentclass[letterpaper, 10 pt, conference]{ieeeconf}  

\IEEEoverridecommandlockouts                              
\renewcommand{\baselinestretch}{0.98}

\usepackage{amsmath} 
\usepackage{amssymb}  

\makeatletter
\let\NAT@parse\undefined
\makeatother

\usepackage{graphicx} 
\usepackage{subcaption}
\usepackage{multirow}
\usepackage[table]{xcolor}
\usepackage{makecell}
\let\labelindent\relax
\usepackage[shortlabels]{enumitem}
\usepackage{algorithm}
\usepackage{algcompatible}
\usepackage{bm}
\usepackage{dcolumn}
\usepackage{dsfont}
\usepackage{adjustbox}
\usepackage{wrapfig}

\usepackage{amsmath,amsfonts,bm}

\def\eqref#1{equation~\ref{#1}}

\def\1{\bm{1}}

\DeclareMathAlphabet{\mathsfit}{\encodingdefault}{\sfdefault}{m}{sl}
\SetMathAlphabet{\mathsfit}{bold}{\encodingdefault}{\sfdefault}{bx}{n}

\def\gA{{\mathcal{A}}}

\def\gR{{\mathcal{R}}}

\newcommand{\E}{\mathbb{E}}

\newcommand{\st}{\mathbf{s}}
\newcommand{\zt}{\mathbf{z}}
\newcommand{\gt}{\mathbf{g}}

\newcommand{\at}{\mathbf{a}}

\usepackage{hyperref}

\title{\LARGE \bf
What Can I Do Here? Learning New Skills \\ by Imagining Visual Affordances
}

\author{Alexander Khazatsky$^{*1}$, Ashvin Nair$^{*1}$, Daniel Jing$^1$, Sergey Levine$^1$
\thanks{$^*$First two authors contributed equally. $^1$UC Berkeley. Correspondence to {\tt\small anair17@berkeley.edu} }
}

\begin{document}

\maketitle
\thispagestyle{empty}
\pagestyle{empty}

\begin{abstract}
A generalist robot equipped with learned skills must be able to perform many tasks in many different environments. However, zero-shot generalization to new settings is not always possible. When the robot encounters a new environment or object, it may need to finetune some of its previously learned skills to accommodate this change. But crucially, previously learned behaviors and models should still be suitable to accelerate this relearning. In this paper, we aim to study how generative models of possible outcomes can allow a robot to learn visual representations of affordances, so that the robot can sample potentially possible outcomes in new situations, and then further train its policy to achieve those outcomes. In effect, prior data is used to learn what kinds of outcomes may be possible, such that when the robot encounters an unfamiliar setting, it can sample potential outcomes from its model, attempt to reach them, and thereby update both its skills and its outcome model. This approach, visuomotor affordance learning (VAL), can be used to train goal-conditioned policies that operate on raw image inputs, and can rapidly learn to manipulate new objects via our proposed affordance-directed exploration scheme. We show that VAL can utilize prior data to solve real-world tasks such drawer opening, grasping, and placing objects in new scenes with only five minutes of online experience in the new scene.
\end{abstract}

\section{Introduction}

Suppose that you need to learn to open a new kind of drawer in a kitchen. While this new skill might demand some amount of trial and error, you would likely be able to use your mental model and past experience to \emph{imagine} the drawer in the open position, and perhaps even imagine likely intermediate steps, such as grasping the handle, even if you do not yet know precisely how to perform the task. Borrowing the terminology put forward by Gibson~\cite{gibson1979ecologicalapproach}, the drawer presents the \emph{affordance} of being ``openable,'' and you are aware of this affordance from your past experience with other similar objects. 
In fact, infants learn about affordances such as movability, suckability, graspability, and digestibility through interaction and sensory feedback~\cite{berger2014development}.
Learning and utilizing affordances through interaction allows an agent to acquire diverse, meaningful experiences even in unfamiliar situations.
However, this way of learning new skills differs markedly from the approach taken by most robotic learning algorithms: the most widely used exploration methods are generally \emph{undirected}, and focus more on seeking out novelty and surprise~\cite{houthooft2016vime, tang2017hashtag, pathak2017curiosity}, rather than familiar and previously seen outcomes.
In this paper, we study how robots operating entirely from pixel input can learn about affordances and, when faced with a new and unfamiliar environment, can utilize a previously trained model of possible outcomes to propose potential \emph{goals} that they can practice in this new environment, so as to explore and update their policy efficiently.

\begin{figure}
  \includegraphics[width=0.99\linewidth]{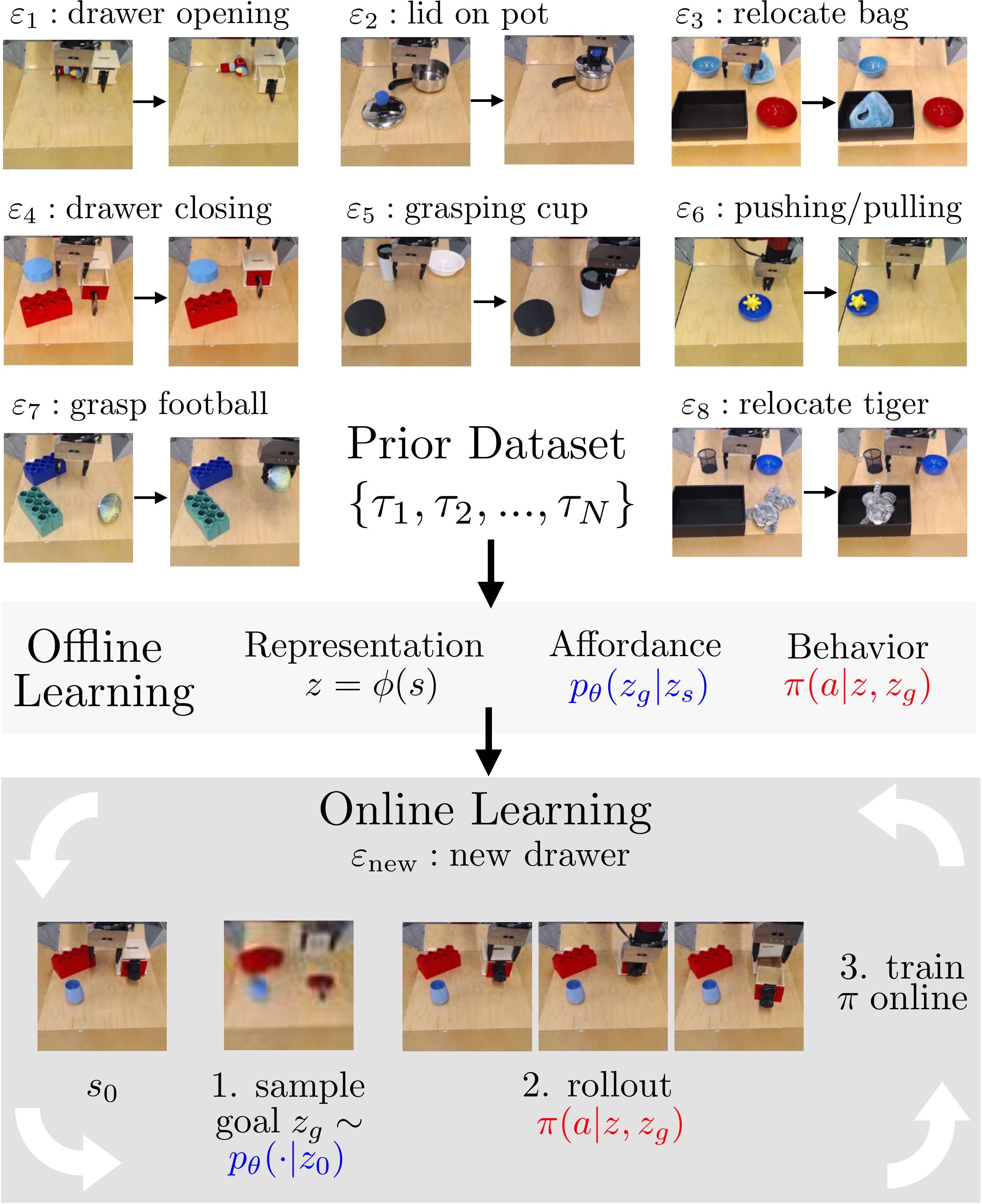}
  \caption{\small
  We propose a system for efficient self-supervised robot learning in an unseen environment $\mathcal{E}_\text{new}$ by utilizing prior data $\mathcal{D}$ of trajectories from related similar environments $\mathcal{E}_i \sim p(\mathcal{E})$. Tasks are specified via a target goal image. From prior data, we learn an encoder $\zt = \phi(\st)$ of images (representation) for compressing observations and self-generating rewards, a model of what tasks might be tested in the new environment (affordance), and goal-conditioned policy to accomplish a given task (behavior). While this provides reasonable performance in some test environments, perfecting the policy in test environments may require additional interaction in the test environment. Dropped in a test environment $\mathcal{E}_\text{new}$ without a given goal, we run RL online in order to practice potential tasks, with goals sampled from the affordance model. Online behavior learning allows us to improve the policy $\pi$ for $\mathcal{E}_\text{new}$ even when it contains new and unseen objects. 
  }
  \label{fig:page1}
  \vspace{-0.5cm}
\end{figure}

We study affordance learning through the framework of self-supervised goal-conditioned reinforcement learning (RL).
Learning a new skill in this framework requires generalization in terms of goal setting (\emph{affordances}) and generalization in terms of goal reaching (\emph{behavior}).
Prior methods for goal-conditioned RL learn a policy to reach a goal state without the need for an externally provided reward function, and are able to master skills such as pushing and grasping objects from image observations~\cite{agrawal2016poking, nair2018rig, lynch2019play, nair2019ccrig}.
They learn skills by setting goals to explore, and learning a policy to reach them.
However, while these prior works have studied how to learn goal-conditioned policies in individual environments, they do not consider what happens when the robot enters a \emph{new} environment where the policy does not simply generalize zero-shot and needs to be trained further. 
Our approach to solving tasks in this setting is to learn affordances, represented by a generative model of possible outcomes, and then sample possible affordances in a new environment to explore the new environment efficiently.

Our method, visuomotor affordance learning (VAL), uses expressive conditional models for learning generalizable affordances, along with off-policy RL to learn generalizable behaviors.
First, we propose to use more expressive generative models for RL to learn compressed representations of images that can reconstruct unseen objects and help the policy generalize to them.
Second, we propose to learn an expressive conditional generative model that generalizes past data to set meaningful goals for novel environments, enabling an understanding of object affordances.
Third, we demonstrate how we can use off-policy RL with all prior data to initialize a general-purpose goal-conditioned policy, then fine-tune the policy with additional data to master a skill in a new environment. 
Combining RL and representation learning, we show that we are able to learn skills with a small amount of online exploration on novel objects in real-world robotic scenarios such as object grasping, relocation, and drawer opening and closing.

The main contribution of this work is to present a learning system that can learn robotic skills entirely from image pixels in novel environments by utilizing prior data to generate goals and generalize behavior in new settings. We demonstrate that our method can learn complex manipulation skills including grasping, drawer opening, pushing, and object re-positioning for a diverse set of objects in simulation. We also demonstrate our method in the real world on a Sawyer robot, where our method is able to learn tasks such as grasping and placing unseen objects and opening and closing unseen drawers after only five minutes of online interaction.

\section{Related Work}
\label{sec:related}
RL has been applied previously to robotic manipulation~\cite{kober2008mp, peters2010reps, levine2016gps}, and also various other applications from playing games~\cite{mnih2013atari, silver2016alphago} to locomotion~\cite{benbrahim1997biped, kohl2004quadruped, deisenroth2011pilco, williams2017mpc}.
Such approaches require an external reward function, but obtaining this reward function itself poses a challenge to exploring in novel uninstrumented environments as we consider in this paper. 
Thus, in this work we focus on self-supervised RL methods that that do not assume externally provided reward functions.

When learning without an externally provided reward function, one common idea is to use novelty-based intrinsic reward functions~\cite{chentanez2005intrinsically, lopes2012exploration, bellemare2016unifying, houthooft2016vime, stadie2016exploration, pathak2017curiosity}.
State novelty-based methods eventually visit all possible states, but do not necessarily learn a useful policy from purely optimizing the exploration objective.
An alternative exploration framework that learns a useful policy even solely from the intrinsic objective is goal reaching~\cite{kaelbling1993goals, schaul2015uva, Baranes2012, andrychowicz2017her, nair2018rig, nachum2018hiro, held2018goalgan, Pere2018, wadefarley2019discern, pong2020skewfit}: by picking a distance measure between states, setting goals, and attempting to reach them, an agent can discover all potential goals in its environment.
We refer the reader to the survey of Colas et al. \cite{colas2021gepsurvey} for a full classification and discussion of these methods.
However, these prior methods do not study the question of how to set goals in new environments, which is vital for collecting coherent experience when faced with a new task.
Our method utilizes representation learning and off-policy RL to generalize prior experience to set exploration goals in new settings.

Most similar to our work, context-conditioned reinforcement learning with imagined goals (CCRIG) learns a conditional variational auto-encoder (CVAE)~\cite{sohn2015cvae} that generates goals conditional on the current scene \cite{nair2019ccrig}. CCRIG was able to learn pushing skills that generalized mainly to object color and partially to object geometry. Our work differs from CCRIG in a number of ways. First, we learn expressive generative models that are able to generate goals in scenes with significantly more visual diversity. Second, we learn a diverse set of skills (e.g., grasping, drawer opening, object placing) that require the goal generation to understand affordances of the environment. Finally, we show that we can use off-policy RL on prior experience, in addition to fine-tuning further on a single specific task to learn new skills. These differences allow our method to better operate in real-world scenarios, as borne out in our experiments.

Another line of work explores the use of affordances in RL, robotics, and control, historically through the lens of perception~\cite{zech2017affordances, hassanin2018affordances, yamanobe2018affordances}. Affordances have also been discussed previously in reinforcement learning in order to accelerate planning in model-based RL by planning over only a subset of relevant actions~\cite{abel2014affordances, khetarpal2020affordances, xu2021affordances}. Our work is more related to the view of affordances in developmental robotics, where affordances were hypothesized to be useful for learning general manipulation skills~\cite{hart2010affordances, min2016affordances}. In our work, we show how goal-conditioned RL utilizing from prior data can put these ideas into practice on real-world robotics systems.

\section{Preliminaries}
\label{sec:background}

In this section, we cover preliminaries on RL, goal-conditioned RL, and self-supervised visual RL.

\noindent \textbf{Goal-conditioned reinforcement learning.}
In goal-conditioned RL, we augment the standard Markov decision process (MDP), which is defined in terms of states $\st_t \in \mathcal{S}$, actions $\at_t \in \mathcal{A}$, and environment dynamics \mbox{$\st_{t+1} \sim p(\cdot|\st_t, \at_t)$}, with goals $\gt \in \mathcal{G}$ that represent the agent's intention to perform one of a variety of tasks drawn from the task family $p(\gt)$. The reward function is also goal-conditioned, and given by some function $r(\st, \gt)$. The discounted return is defined as $R_t = \sum_{i=0}^H \gamma^i r(\st_i, \gt)$, where $\gamma$ is a discount factor and $H$ is the horizon, which may be infinite. The aim of the agent is to optimize a policy $\pi(\at_t|\st_t, \gt)$ to maximize the expected discounted return $J(\pi) = \E_\gt[R_0]$.
Efficient off-policy RL algorithms have been proposed to learn goal-conditioned policies~\cite{schaul2015uva, andrychowicz2017her}.

\noindent \textbf{Self-supervised visual reinforcement learning.}
For scalable robot learning, we cannot always assume known shaped reward functions for tasks. In the absence of such reward functions, Andrychowicz et al. propose goal-state reaching as a natural objective~\cite{andrychowicz2017her}:
tasks are defined by state outcomes, where the goal space $\mathcal{G} = \mathcal{S}$, the state space, and the reward is a goal-reaching objective $r(\st, \gt) = -\mathds{1}_{||\st - \gt|| > \epsilon}$.
The task distribution $p(\gt)$ is chosen to be the feasible states of the robot and objects it interacts with.

But when states are high-dimensional (e.g., images), two issues arise: we do not know the task distribution $p(\gt)$, and exact reaching of a target goal state is impractical.
Reinforcement learning with imagined goals (RIG)~\cite{nair2018rig} addresses these issues with a generative model, which is used to learn a latent space of observations and similarity metric on images. 
Specifically, a variational auto-encoder~\cite{kingma2014vae} with encoder $\phi(\zt_t|\st_t)$ and prior $p(\zt)$ is learned. 
At training time, the robot sets goals for itself in latent space by sampling a goal latent $\zt_g \sim p(\zt)$ and learns a policy $\pi(\zt_t, \zt_g)$ to reach latent goals. 
At test time, the robot can be tasked with a goal image $\gt$ and execute the learned policy with goal latent $\zt_g \sim \phi(\cdot|\gt)$ to match the goal image.
In this way, goal-conditioned RL with generative models enables self-supervised learning in a single environment $\mathcal{E}$ where the task distribution $p(\gt)$ is not known apriori.

\section{Problem Setting}

In this paper, we now consider acting in a distribution of environments $p(\mathcal{E})$ with shared structure.\footnote{Meta-learning approaches have also studied learning a new task quickly, given experience on a set of related MDPs ~\cite{duan2016rl2, finn2017maml, rakelly2019pearl}. The aims of our method are related to meta-learning, in that we also aim to learn in new environments more quickly, but we do not assume being given user-specified tasks or rewards to solve in the new environment.}
Each environment $\mathcal{E}_i$ has its own task distribution $p_i(\gt)$. 
As before, tasks are defined by goal states, so $p_i(\gt)$ represents the potential outcomes of interest in that environment.
We assume that the outcomes of interest depends only on the appearance of the environment.
When $\mathcal{E}$ is fully observed, this means $p(\gt|s_0)$ is shared across environments.
As prior data, the robot has access to a training dataset $\mathcal{D} = \{\tau_1, \tau_2, \dots \tau_N\}$ of trajectories from prior environments, where the trajectories achieve a final outcome $\st_T \sim p_i(\gt)$ -- that is, they succeed on tasks from the underlying task distribution for that environment. 
Now, the robot is placed in a new environment $\mathcal{E}_\text{new} \sim p(\mathcal{E})$, and must learn to solve tasks in the new environment through self-supervised practice at training time, such that it can accomplish tasks sampled from $p_i(\gt)$ at test time.
The environments we consider vary visually and dynamically in terms of the objects that are present and potential tasks they afford, such as being able to lift various objects and open different drawers; such real-world variation is presented in Figure~\ref{fig:page1}.

Thus, the agent in this new setting must generalize its prior experience $\mathcal{D}$ to practice potential skills it may be asked to perform at test time in the new environment efficiently, even when it encounters novel objects.
At test time, the robot is evaluated in terms of its ability to accomplish a task in $\mathcal{E}_\text{new}$ specified by a goal image.
To perform well in this setting, a method must attempt to infer $p_\text{new}(\gt)$, which should be possible since $p(\gt|s_0)$ is common across environments and $\mathcal{D}$ contains trajectories that achieve goals from the same distribution.
Note that given observations from $\mathcal{E}_\text{new}$, the agent may still have to practice the various behaviors the environment affords if there are multiple, since the test task distribution is unknown at training time.

\section{Visuomotor Affordance Learning}
\label{sec:method}

In this section, we present visuomotor affordance learning (VAL), our method for self-supervised learning in novel environments utilizing prior data from related environments. VAL consists of three learning phases: (A) an affordance learning phase to learn affordances from the prior data, (B) an offline behavior learning phase to learn behaviors from the prior data, and (C) an online behavior learning phase where the agent actively interacts with the test environment using affordances and learns potential behaviors in the new environment.
The overall method is summarized in Algorithm~\ref{algo:val}.

\begin{algorithm}[t]
\caption{Visual Affordance Learning}
\label{algo:val}
\begin{algorithmic}[1]
\REQUIRE Dataset $\mathcal{D}$, policy $\pi(\at|\zt, \zt_g)$, $Q$-function $Q(\zt, \at, \zt_g)$, RL algorithm $\gA$, replay buffer $\gR$, relabeling strategy
$p_\text{RS}(\zt)$, environment family $p(\mathcal{E})$.
\STATE Learn encoder $\phi(\zt|\st)$ by generative model of $\mathcal{D}$
\STATE Learn affordances $p(\zt_t|\zt_0)$ by generative model of $\mathcal{D}$
\STATE Add latent encoding of $\mathcal{D}$ to the replay buffer
\STATE Initialize $\pi$ and $Q$ by running $\gA$ offline
\STATE Sample $\mathcal{E}_\text{new} \sim p(\mathcal{E})$, $\mathcal{E}_\text{new} = \left(p_\text{new}(\st_0), \pi_\text{new}(\st_{t+1} | \st, \at)\right)$
\FOR{$1, \dots, N_\text{episodes}$}
    \STATE Sample initial state $\st_0 \sim p_\text{new}(\st_0)$.
    \STATE Sample goal $\zt_g \sim p(\zt_t|\zt_0)$
    \FOR{$t = 0, \dots, H$}
        \STATE Sample $\at_t \sim \pi(\cdot|\zt_t, \zt_g)$
        \STATE Sample $\st_{t+1} \sim p_\text{new}(\cdot | \st_t, \at_t)$
    \ENDFOR
    \STATE Store trajectory $(\zt_1, \at_1, \dots, \zt_H)$ in replay buffer $\gR$.
    \FOR{$1, \dots, N_\text{train\_steps}$}
        \STATE Sample transition $(\zt_t, \at_t, \zt_{t+1}, \zt_g)$
        \STATE Relabel with $\zt_g' \sim p_\text{RS}(\zt_g)$ and recompute reward
        \STATE Update $\pi$ and $Q$ with relabeled transition using $\gA$
    \ENDFOR
\ENDFOR
\end{algorithmic}
\end{algorithm}

\subsection{Affordance Learning}
\label{sec:method_affordance_learning}

The affordance learning system must generate goals that induce coherent exploration trajectories even in new environments.
We would like to learn a model that, given an observation $s_0$ in a new environment, generates a potential goal state the agent might be tasked to reach.
With high-dimensional image observations, attempting to sample such goal states directly is a difficult generative modeling problem.
Instead, we first learn a lower-dimensional latent space $p(\zt_t|\st_t)$ by training a generative model through image reconstruction.
With sufficiently expressive models and enough data, the generative model represents images in a manner that it can reconstruct even unseen objects.
Given such a latent space, we can then learn affordances by training a conditional model $p(\zt_t|\zt_0)$
to generate goals that are plausible outcomes of an initial state, even for unseen environments.

To instantiate these two models, we must choose a class of latent variable generative models for $p(\st_t | \zt_t)$ and conditional affordance models $p(\zt_t | \zt_0)$. For the first, while many choices would be suitable, including models such as variational auto-encoders (VAEs)~\cite{kingma2014vae} and generative adversarial networks (GANs)~\cite{goodfellow2014gan, donahue2017bigan}, in our implementation we use the VQVAE model~\cite{oord2017vqvae}. The VQVAE is expressive enough to represent very diverse datasets, and be able to reconstruct even unseen objects with a high level of detail. We do not require image reconstructions for our method, but the ability to partially reconstruct unseen objects suggests that model expressively represents geometry and color information that may be important for learning affordances and behaviors. In the VQVAE case, $\phi$ is deterministic, so we will use the shorthand for the latent embedding $\zt_t = \phi(\st_t)$, where $\zt_t$ is the continuous latent resulting after quantization. In our experiments, we compare this choice of model to other expressive models: VAE~\cite{kingma2014vae}, CVAE~\cite{sohn2015cvae}, and BiGAN~\cite{donahue2017bigan}.

Next, given a latent space, we need to sample potential goals from this space that is predictive of which goals might be tasked at test time in the new environment $\mathcal{E}_\text{new}$. We use a conditional PixelCNN model~\cite{oord2016pixelcnn} in the latent space to do so, conditioned on the initial state $s_0$. The PixelCNN model is trained to maximize $\E_{\tau \sim \mathcal{D}, (\st_0, \st_t) \sim \tau, z \sim \phi(\cdot|\st)}[\log p_\theta(\zt_t|\zt_0)]$, where $\theta$ is the parameters of the PixelCNN density model. To generate exploration goals, we sample $z_g \sim p_\theta(\cdot|\zt_0)$, where $\zt_0 = \phi(\st_0)$ is the encoding of an image of the current setting. The PixelCNN model in VAL being conditional allows us to generate meaningful goals that might be achievable with a novel object. As we show in Section~\ref{sec:gen_model_results}, the conditional model allows us to sample affordances such as opening drawers and lifting objects, even for new environments that are visually complex with variation in the identity, color, geometry, and functionality of objects.

\subsection{Offline Behavior Learning}
\label{sec:method_behavior_learning}
Given learned affordances of what behaviors the agent may be tasked with, the agent needs to learn how to actually accomplish those behaviors. In this phase, we wish to learn a reasonable goal-conditioned policy with offline RL. While trained on a limited fixed offline dataset, the hope is that the policy can generalize learning from offline data to accomplish desired behaviors in a new environment, allowing us to either perform tasks successfully zero-shot without further learning, or collect meaningful exploration data in the next phase (Section \ref{sec:online_policy}).

To learn with offline RL while allowing the possibility of quickly fine-tuning in a new environment, we use advantage weighted actor critic (AWAC)~\cite{nair2020awac} as the underlying RL algorithm.
AWAC is an off-policy RL algorithm that has shown strong performance in utilizing prior data for offline pretraining while still being amenable to online fine-tuning.
The aim of behavior learning is to optimize a goal-conditioned policy $\pi(\at|\zt, \zt_g)$ which is able to solve any task in the task distribution $p(\zt_g)$.
We do not assume an external reward function, so we require a reward function to optimize. 
Following prior work~\cite{andrychowicz2017her, nair2018rig}, we optimize a goal reaching objective: to maximize the density $p_{\mathbf{Z}_g}(\zt = \zt_g)$; in practice we use the sparse reward function $r(\zt, \zt_g) = -\mathds{1}_{||\zt - \zt_g|| > \epsilon}$, where $\epsilon$ is a fixed threshold as it encourages fully solving tasks in a binary fashion.
For a particular transition in the replay buffer $(\zt_t, \at_t, \zt_{t+1}, \zt_g, r)$, we can also  relabel the goal with a new goal $\zt_g'$ and the recompute the reward. In practice we keep $\zt_g$ with 20\% probability, future hindsight experience replay with 40\% probability, and sample $\zt_g' \sim p(z_t|z_0)$ with 40\% probability.
Importantly, due to using a compressed representation, we can expect that $z_t$ in any new environment will be semantically similar to past experience. Thus, in this phase, we can obtain a policy that generalizes partially to a new environment.

\begin{figure}[t]
    \begin{subfigure}[b]{0.14\textwidth}
        \begin{flushright}
        
        Initial Image
        
        \vspace{0.5cm}
        
        CVQVAE (Ours)
        
        \vspace{0.5cm}
        
        CCVAE
        
        \vspace{0.35cm}
        \end{flushright}
    \end{subfigure}
    \begin{subfigure}[b]{0.3\textwidth}
        \scalebox{1.8}{
            \includegraphics[width=0.1\textwidth]{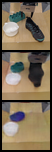} 
            \includegraphics[width=0.1\textwidth]{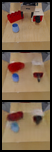}
            \includegraphics[width=0.1\textwidth]{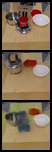}
            \includegraphics[width=0.1\textwidth]{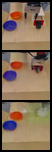}
            \includegraphics[width=0.1\textwidth]{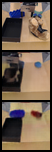}
        }
    \end{subfigure}

    \caption{Samples on unseen objects in the real world. In each column, the top image is the conditioning image $s_0$ and the images below are conditionally sampled images from the corresponding generative model. Our model, CVQVAE, generates clear and diverse samples. }
    \label{fig:real_samples}
    \vspace{-0.5cm}
\end{figure}

\subsection{Online Behavior Learning}
\label{sec:online_policy}

Guaranteeing zero-shot generalization for every possible new environment is in general impossible. Instead, we will discuss how to finetune in a specific environment $\mathcal{E}_\text{new}$ using affordances. 
In this phase, we utilize the learned affordances which inform \textit{what} tasks to perform, and the offline learned behaviors that inform \textit{how} to perform those tasks.

At online training time, the new task distribution $p_\text{new}(\gt)$ is unknown, so we use the affordance model to sample potential tasks.
Thus, to collect coherent exploration data, we sample goals from the affordance module $z_g \sim p_\theta(\cdot|z_0)$ and roll out the goal-conditioned policy $\pi(\at|\zt, \zt_g)$.
We then iterate between improving the policy with off-policy RL, and collecting exploration data and appending it to the replay buffer.
To learn a new task, exploration data in the new environment for fine-tuning the policy is extremely valuable, so this iteration allows us to quickly fine-tune. The online learning process is illustrated in the bottom box in Figure~\ref{fig:page1}.

In summary, VAL enables self-supervised learning in novel environments utilizing prior data. 
We first learn a generative model for learning a latent space and affordances, and learn how to accomplish tasks with off-policy RL. 
Then, both are used in an online behavior learning phase to perfect potential behaviors in a new environment. 
The overall method is summarized in Algorithm~\ref{algo:val}.

\begin{figure*}[t]
    \renewcommand{\arraystretch}{1.5}
    \begin{subfigure}[t]{0.6\textwidth}
        \raisebox{2.1cm}{
        
        \begin{tabular}{ l||m{2.8cm}|m{2.5cm}  }
            \centering
            Task & VAL (Ours) \; \; \; \; Offline $\rightarrow$ Online & CCRIG  \; \; \; \; \; \; Offline $\rightarrow$ Online \\
            \hline
            (1) Pickup shoe & $12.5\%$ $\rightarrow$ $\textbf{50\%}$ & $0\%$ $\rightarrow$ $16.6\%$ \\
            (2) Drawer closing & $25\%$ $\rightarrow$ $\textbf{100\%}$ & $0\%$ $\rightarrow$ $12.5\%$ \\
            (3) Drawer opening & $62.5\%$ $\rightarrow$ $\textbf{100\%}$ & $0\%$ $\rightarrow$ $0\%$ \\
            (4) Place object in tray & $25\%$ $\rightarrow$ $\textbf{75\%}$ & $0\%$ $\rightarrow$ $0\%$ \\
            (5) Lid on pot & $37.5\%$ $\rightarrow$ $\textbf{87.5\%}$ & $0\%$ $\rightarrow$ $0\%$ \\
        \end{tabular}
        }
    \end{subfigure}
    \begin{subfigure}[t]{0.05\textwidth}
        \vspace{-3.3cm}
        (1)
        \vspace{0.3cm}
        
        (2)
        \vspace{0.3cm}
        
        (3)
        \vspace{0.3cm}
        
        (4)
        \vspace{0.3cm}
        
        (5)
    \end{subfigure}
    \begin{subfigure}[t]{0.34\textwidth}
        \includegraphics[width=0.99\textwidth]{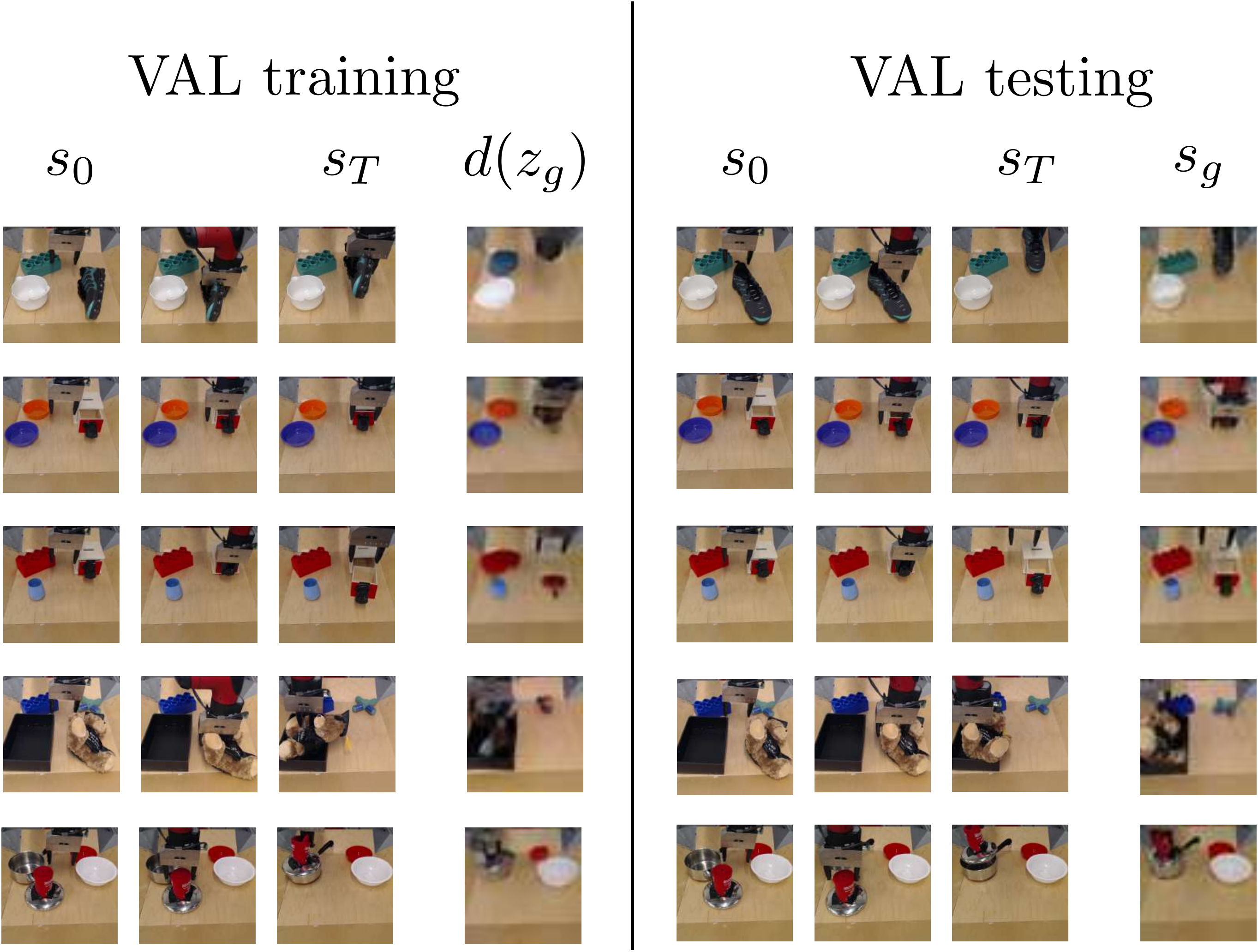}
    \end{subfigure}
    
    \caption{Real-world results. Left, success rates per method for the five tasks tested. We report the offline performance, followed by the performance after five minutes of online fine-tuning. With VAL, we see reasonable initial offline performance followed by significant improvement on all tasks. Meanwhile CCRIG fails to succeed any of the tasks. Right, film strips of VAL during training (left, with decoded affordance proposals) and testing (right, with goal images) are visualized. Videos are available at \url{https://sites.google.com/view/val-rl}}
    \label{fig:real-world-results}
\end{figure*}

\section{Real-World Experimental Evaluation}
\label{sec:result}

We first evaluate our method in a real-world manipulation setting with a Sawyer robot and diverse objects and tasks. This setting is very challenging due to the variety of objects, scenes, and tasks that the robot interacts with, and being limited to using image inputs.

\textbf{Real-world setting.} The real-world setting is shown in Figure~\ref{fig:page1}. A Sawyer robot is controlled at 5Hz with 4 degrees of control: 3 dimensions of end-effector velocity control in Euclidean space and one dimension to open and close the gripper. The environments span 10 drawer handles, 10 pot handles, 40 toys, and 60 distractor objects. To create a new environment $\mathcal{E}$ in the real world, we randomly sample one interaction object as well as 2-3 distractor objects. The behaviors that the environments afford therefore span grasping and relocating toys, opening and closing drawers, as well as covering and uncovering pots. Each test environment contains an unseen interaction object and a random set of 2-3 distractor objects, with all positions randomized.

Our experiments aim to answer the following questions:
\begin{enumerate}
    \item Does the learned generative model sample plausible affordances in new scenes?
    \item Is VAL able to accelerate learning of real-world manipulation skills online in new settings?
\end{enumerate}

\subsection{Generative Models for Affordance Learning}
\label{sec:gen_model_results}

Expressive generative models enable us to propose desired outcomes in new environments even when the current policy cannot yet achieve them.
We can preview this capability by inspecting the training procedure of the models,
which also gives insights into which models will tend to perform well when used for interactive learning.
Specifically, we inspect model samples to see if the model can actually output plausible candidate affordances.
We evaluate our model, which combines a VQVAE with a conditional pixel CNN, and prior models that have been used in self-supervised RL as well as other expressive generative models.
We compare (1) VAE~\cite{kingma2014vae}, (2) CVAE~\cite{sohn2015cvae}, (3) BiGAN~\cite{donahue2017bigan}, (4) Conditional BiGAN, (5) Ours, a VQVAE with a conditional PixelCNN.

Sampled affordances are shown in Figure~\ref{fig:real_samples}. Our model is able to sample coherent tasks to perform where other methods produce indistinct or uninterruptible samples. Most interestingly, the VQVAE model turns an unseen drawer (with a different color and handle) into a red drawer, in open and closed positions. This  illustrates its ability to utilize prior data for generating conducive goals for online RL. CCVAE samples are usually less coherent, often missing the object completely and not capturing the geometry of unseen objects. The CBiGAN also struggles to produce realistic images, while the VAE fails completely as it is not a conditional model cannot represent the wide diversity of potential images.

\subsection{Real-World Visuomotor Affordance Learning}

Next, we investigate whether VAL can handle real-world visual complexity and object diversity to learn control policies for varied tasks on a Sawyer robot.
The setup is shown in Figure~\ref{fig:page1}: the robot is tasked with fine-tuning in a particular environment, beginning with about 1,000 trajectories of prior data for affordance learning and offline behavior learning.
The protocol for collecting prior data, as well as examples of images from the data are shown in Appendix~\ref{sec:appendix_real}.

After pretraining affordances, a policy, and a Q function on the prior data, the robot is dropped in a new test environment. We evaluate five test environments which each contain distractor objects as well as objects that may be interacted with such a shoe that can be picked up, drawers than can be opened or closed, and a lid which may be placed on a pot. These behaviors are demonstrated on similar objects in the prior dataset, but the objects during test time are previously unseen - for instance, the drawer has a different handle. The agent can interact with the environment without supervision to collect more data and improve the policy; then to evaluate, the agent is tasked with a specific goal image that corresponds to a task such as opening a closed drawer.

\begin{figure*}[t]
    \centering
    \begin{subfigure}[b]{0.222\textwidth}
        \center
        \includegraphics[width=3cm]{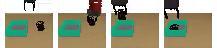}
        
        \vspace{0.1cm}
        \includegraphics[width=0.95\textwidth]{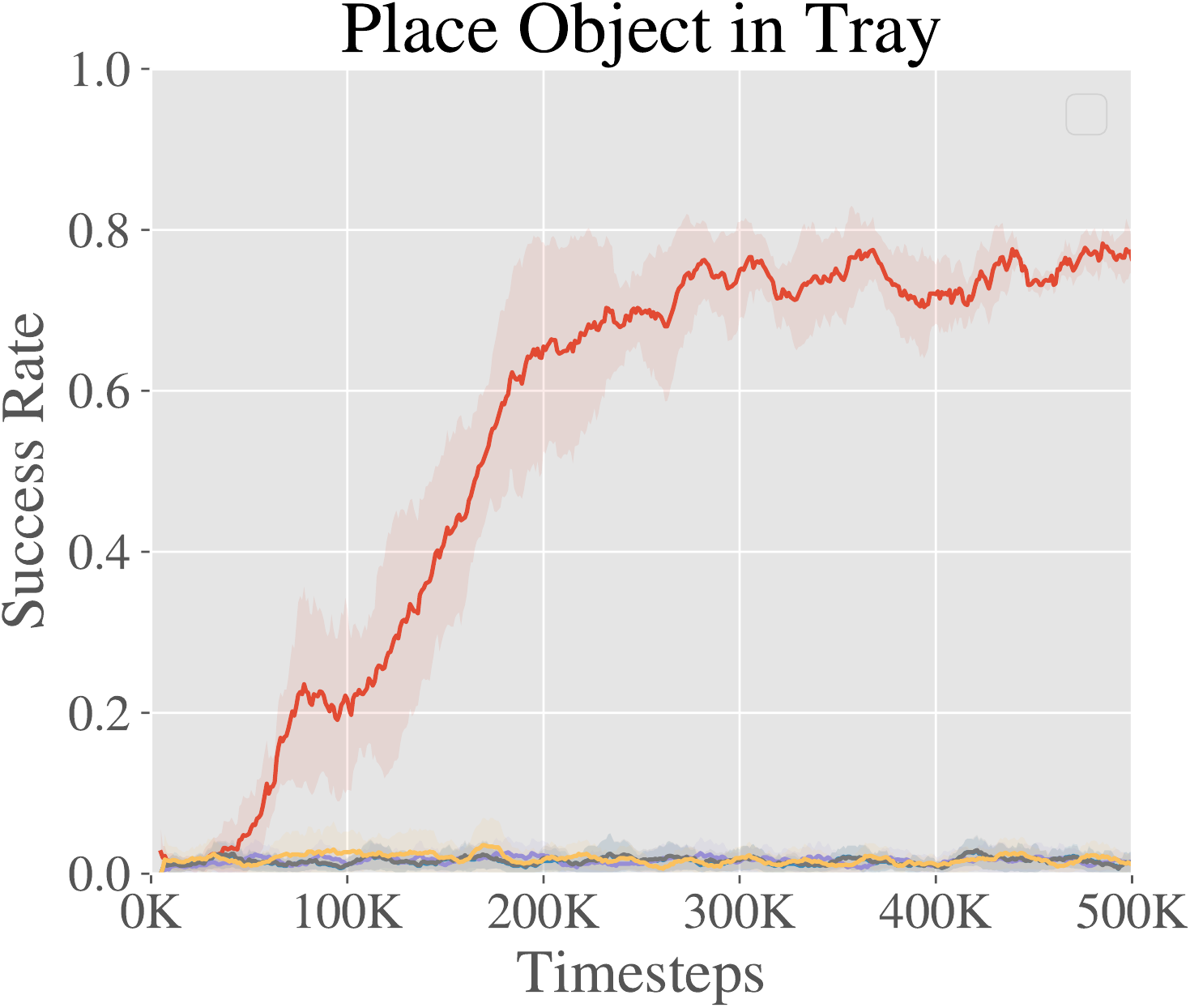}
    \end{subfigure}
    \begin{subfigure}[b]{0.21\textwidth}
        \center
        \includegraphics[width=3cm]{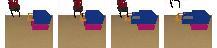}
        
        \vspace{0.1cm}
        \includegraphics[width=1\textwidth]{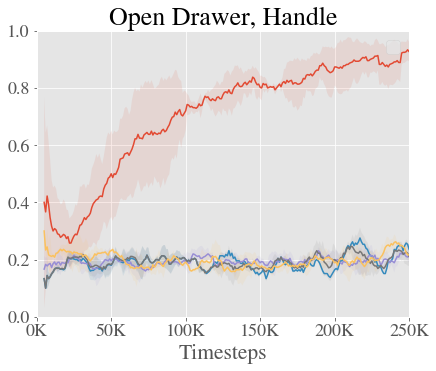}

    \end{subfigure}
    \begin{subfigure}[b]{0.21\textwidth}
        \center
        \includegraphics[width=3cm]{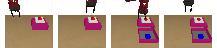}
        
        \vspace{0.1cm}
        \includegraphics[width=1\textwidth]{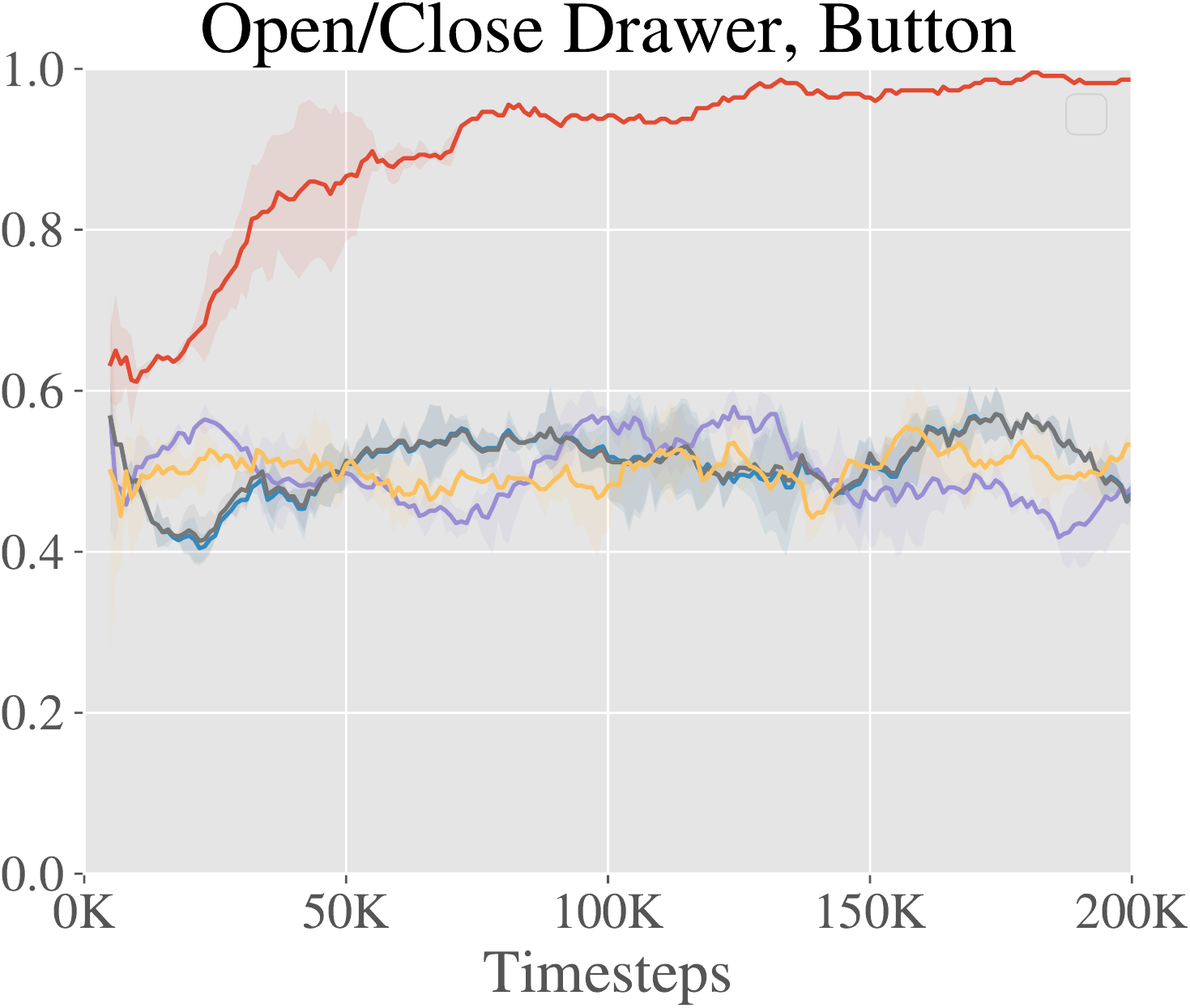}
    \end{subfigure}
    \begin{subfigure}[b]{0.21\textwidth}
        \center
        \includegraphics[width=3cm]{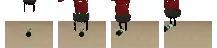}
        
        \vspace{0.1cm}
        \includegraphics[width=1\textwidth]{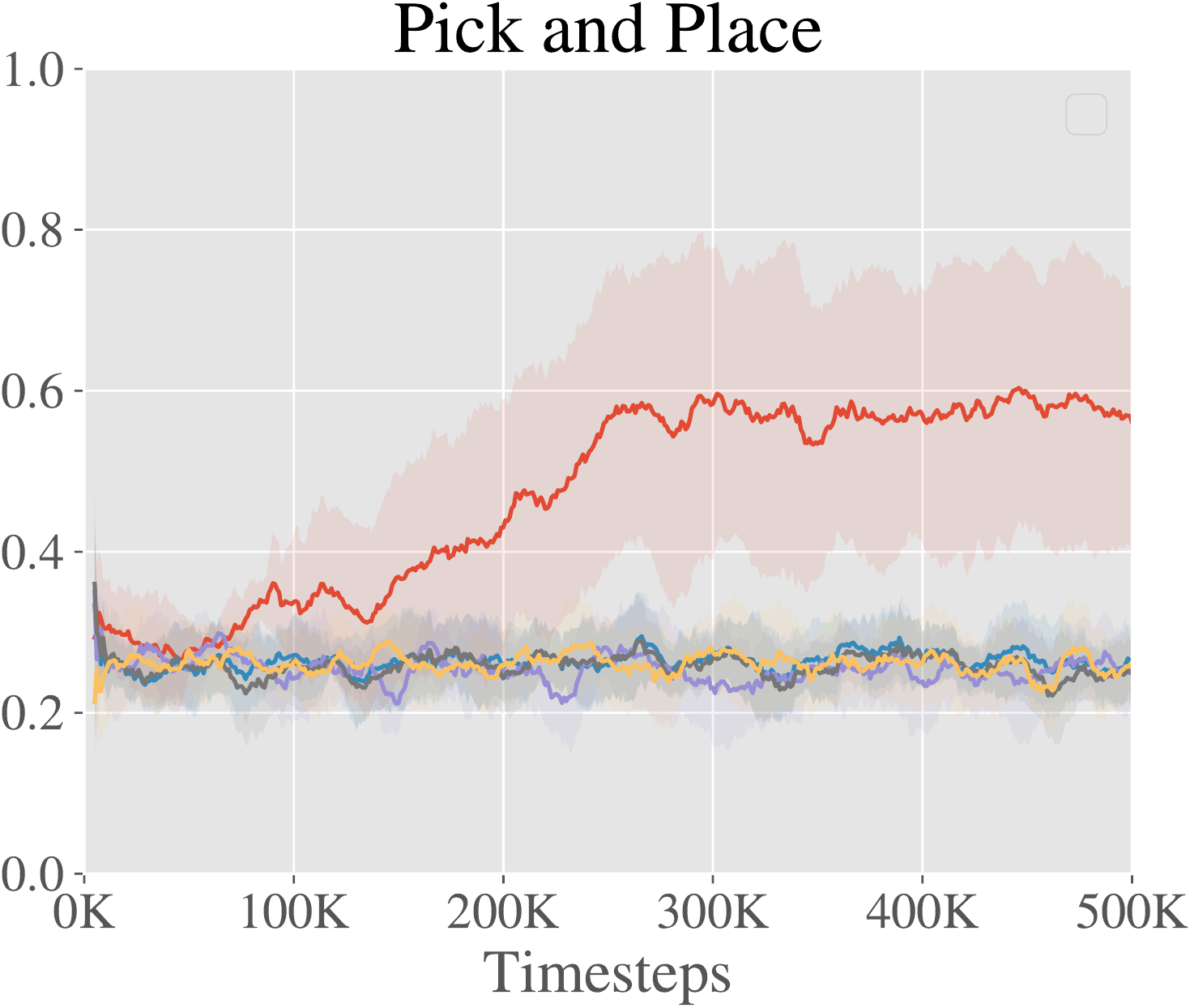}
    \end{subfigure}
    \begin{subfigure}[b]{0.1\textwidth}
        \center
        \includegraphics[width=1\textwidth]{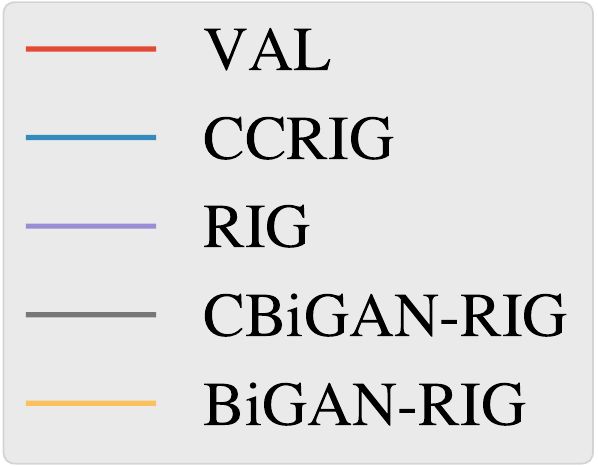}
        \vspace{0.6cm}
    \end{subfigure}
    
    \caption{Learning curves for simulation experiments, fine-tuning on an unseen environment. Our method is able to learn these tasks online, while none of the baselines or prior methods are able to make meaningful learning progress in this setting. A successful rollout of each task in a test environment is shown above the corresponding plots. }
    \label{fig:sim_comparison}
    \vspace{-0.5cm}
\end{figure*}

\begin{figure}[t]
    \center
    \includegraphics[height=0.24\linewidth]{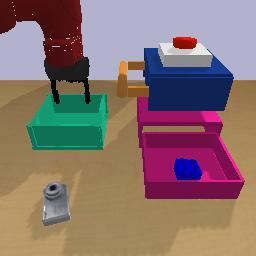}
    \includegraphics[width=0.24\linewidth]{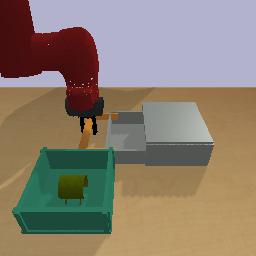}
    \includegraphics[height=0.24\linewidth]{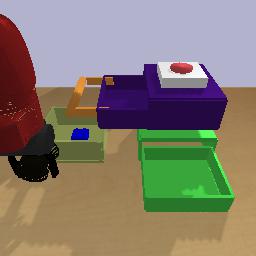}
    \includegraphics[width=0.24\linewidth]{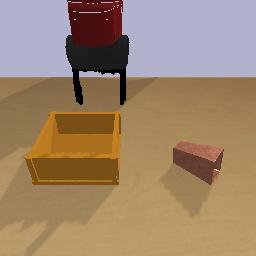}
    \caption{Randomly sampled scenes from our simulated multi-task environment. To practice in a sampled scene, the agent must infer the potential behaviors that the scene affords.}
    \label{fig:simenvs}
\end{figure}

\begin{figure}[t]
    \center
    \includegraphics[width=0.99\linewidth]{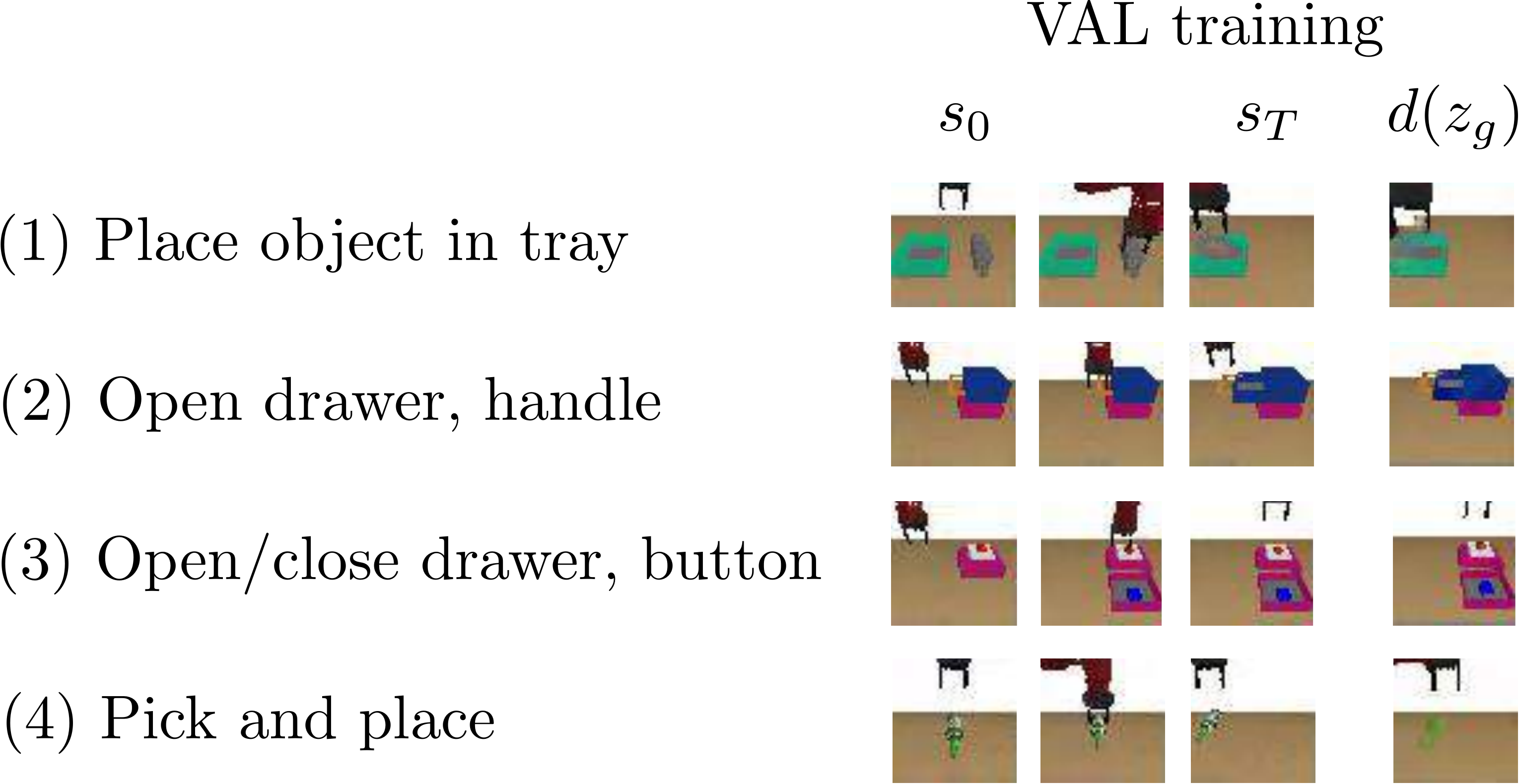}
    \caption{Training rollouts from VAL. For each environment, the reconstruction of a sampled affordance is shown on the rightmost column, and frames from a trajectory attempting to achieve that affordance is shown on the left.}
    \label{fig:simenvs}
\end{figure}

\begin{figure}[t]
    \begin{subfigure}[b]{0.14\textwidth}
        \begin{flushright}
        
        Initial Image
        
        \vspace{0.35cm}
        
        CVQVAE (Ours)
        
        \vspace{0.35cm}
        
        CCVAE
        
        \vspace{0.35cm}
        
        CBiGAN
        
        \vspace{0.35cm}
        
        VAE
        
        \vspace{0.22cm}
        
        \end{flushright}
        
    \end{subfigure}
    \begin{subfigure}[b]{0.3\textwidth}
        \center
        
        \scalebox{1.5}{
                \adjustbox{trim=0 16 0 0,clip} { \includegraphics[width=0.1\textwidth]{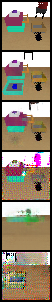} } \hspace{-12px}
                \adjustbox{trim=0 16 0 0,clip} { \includegraphics[width=0.1\textwidth]{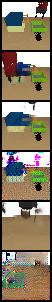} } \hspace{-12px}
                \adjustbox{trim=0 16 0 0,clip} { \includegraphics[width=0.1\textwidth]{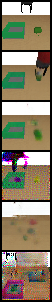} } \hspace{-12px}
                \adjustbox{trim=0 16 0 0,clip} { \includegraphics[width=0.1\textwidth]{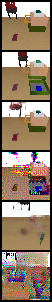} } \hspace{-12px}
                \adjustbox{trim=0 16 0 0,clip} { \includegraphics[width=0.1\textwidth]{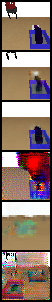} } \hspace{-12px}
                \adjustbox{trim=0 16 0 0,clip} { \includegraphics[width=0.1\textwidth]{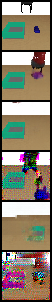} } \hspace{-12px}
            }
    \end{subfigure}
    \caption{Samples on test environments in simulation. In each column, the top image is the conditioning image $s_0$ and the images below are conditionally sampled images. The left four columns are relatively successful samples for our affordance model, each showing the potential outcome of a behavior. The right two columns are failure modes, lacking diversity or altering an object's geometry or color. }
    \label{fig:samples}
\end{figure}

Results running VAL in the real world are shown in the table in Figure~\ref{fig:real-world-results}, reporting the success rates on five test tasks for our method and CCRIG both offline and after one epoch of online training, which includes 10 interactive trials, amounting to less than five minutes of real-world interaction time. On all five tasks, VAL shows nontrivial offline performance followed by strong online improvement. Qualitatively, our method is able to learn behaviors such as recovering from a missed grasp by returning to the grasp. Film strips of our method are shown on the right side of Figure~\ref{fig:real-world-results}. In contrast, CCRIG struggles to make progress on all five tasks. CCRIG fails for two reasons. First, the quality of the sampled affordances are significantly poorer. Second, as can be seen in videos, CCRIG sometimes comes close to solving the task, but does not fully solve it, preventing improvement after subsequent training.

\section{Experimental Evaluation in Simulation}

In order to further study and understand VAL, we carry out more experiments in simulation, where we can better control the quantity of data and ablate parts of the method. These experiments aim to answer the following questions:
\begin{enumerate}
    \item Does VAL outperform prior self-supervised RL methods in accelerating learning a new task from prior data?
    \item Can VAL scale with additional data for affordance and behavior learning?
\end{enumerate}

\textbf{Simulated setting.} Randomly generated workspaces in our simulated multi-task environment are shown in Figure~\ref{fig:samples}. In this PyBullet simulation~\cite{coumans2021}, the robot is faced with multiple potential tasks in each environment based on which objects are present: opening and closing a drawer by the handle, opening and closing a different drawer by pressing a button, grasping objects, and moving objects into drawers or a box. To sample a new environment $\mathcal{E}$, we randomize the existence, position, color, and orientation of the following: two drawers, a box, a button, and an object. If an object is present it is chosen from a set of 84 object geometries. To successfully learn in a test environment, the method must be able to explore in the environment based on the behaviors that environment affords.

\subsection{Self-Supervised Online Fine-Tuning from Prior Data}
We first compare VAL against prior methods in this simulated setting.
The robot receives a prior dataset $\mathcal{D}$ of trajectories in the pre-training environments, which is utilized for learning affordances and offline RL. The details of this dataset are explained further in Appendix~\ref{sec:appendix_sim}. After the pre-training phase, the robot is placed in a test setting and begins online fine-tuning. We evaluate the policy on goal images in the new environment, sampled from expert trajectories, and report whether the final state of the policy's trajectory matches the state of the goal image within a chosen threshold.

Learning curves of the online training phase for various objects are shown in Figure~\ref{fig:sim_comparison}.
We see that for each task, VAL outperforms prior methods which do not make progress on learning these tasks at all.
On the drawer opening task, offline RL achieves nontrivial performance of about 60\%, but online interaction allows fine-tuning to over 90\% success rate.

How is VAL able to outperform prior methods so significantly? One major reason is the quality of sampled goals.
Samples for the affordance model for the simulated domain are shown in Figure~\ref{fig:samples}. 
We compare (1) VAE~\cite{kingma2014vae}, (2) CVAE~\cite{sohn2015cvae}, (3) BiGAN~\cite{donahue2017bigan}, (4) Conditional BiGAN, (5) Ours, a VQVAE with a conditional PixelCNN.
Our model produces diverse, coherent samples of possible outcomes (i.e., affordances) in new scenes with novel objects.
In comparison to our model, conditional VAE samples tend to be blurry and do not capture the geometry of unseen objects well.

\subsection{Scalable Robot Learning with VAL}
For general-purpose robot learning, we would ideally like to learn diverse skills continuously, using prior experience to perpetually improve at learning new skills.
To study this possibility, we examine whether solving tasks in a new environment can be sped up by collecting larger amounts of prior experience.
In this experiment conducted in simulation in order to carefully control the data quantity, the robot receives a prior dataset $\mathcal{D}$ of $K$ trajectories for running VAL to grasp an unseen object in a new environment.
We vary $K$ to observe whether the method can benefit from larger amounts of prior data.

\begin{figure}[t]
    \centering
    \begin{subfigure}[b]{0.99\linewidth}
        \center
        \includegraphics[width=0.6\textwidth]{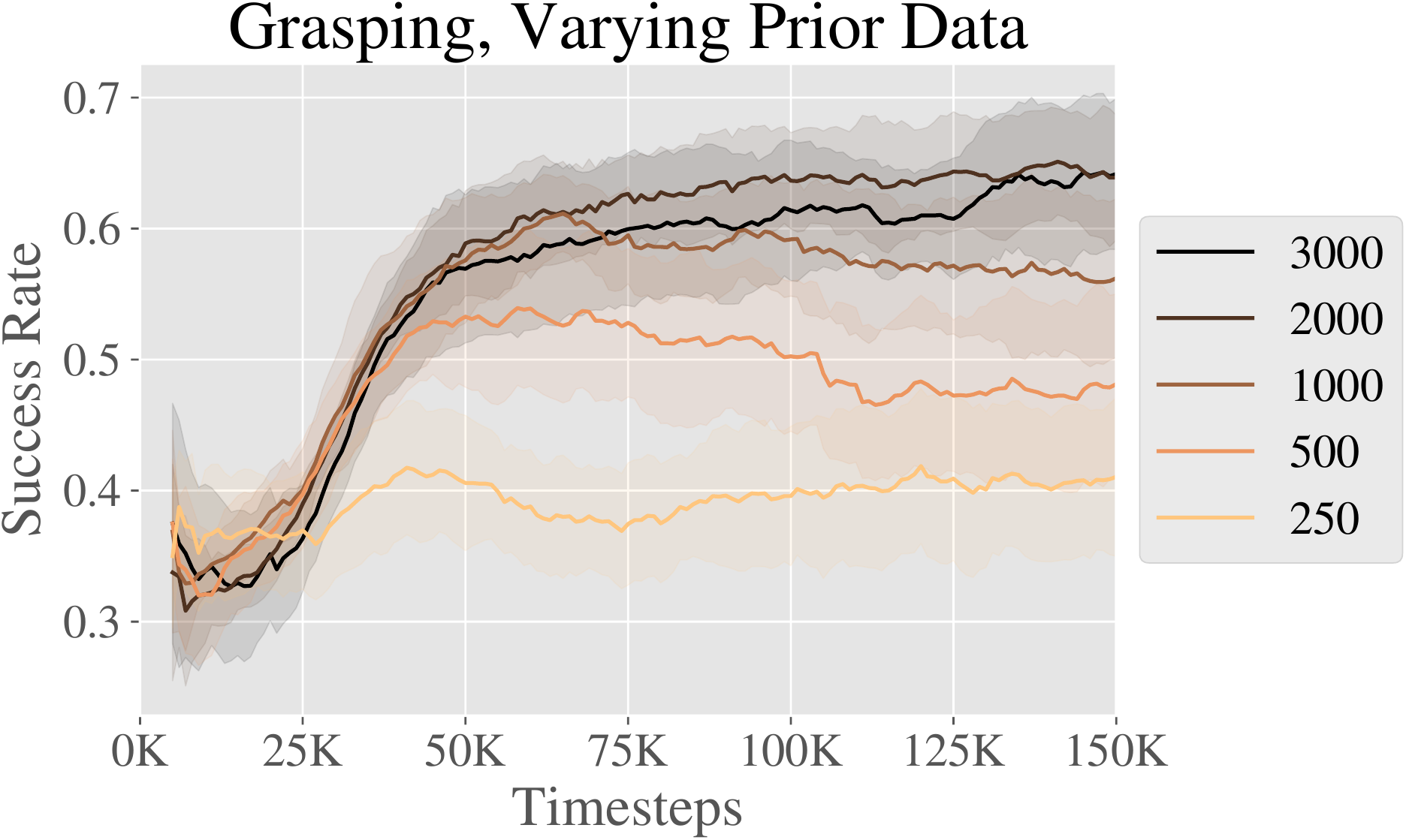}
    \end{subfigure}

    \caption{Learning curves for simulated grasping of novel objects with VAL, using data from an increasing number of training objects for offline RL. We collect 50 trajectories per training object, and each line is labeled with the total number of training trajectories. }

    \label{fig:novel_obj}
    \vspace{-0.5cm}
\end{figure}

Learning curves of the online training phase, averaged over five test objects, are shown in Figure~\ref{fig:novel_obj}. 
First, we see from the starting point of each curve that the offline policy already generalizes to some level for grasping objects, but the average success rate is only around 35\%. 
Importantly, note that training on more data only slightly improves generalization after offline training (at timestep 0).
Then, fine-tuning results in rapid policy improvement up to around 65\% success rate after only 150,000 timesteps when utilizing the most prior data, compared to 40\% with the least prior data.
Thus, more prior data significantly accelerates learning in new environments even when the initial performance is comparable.
This suggests that VAL can be deployed in a continual learning setting, with each new task being learned faster as it benefits from the increasing dataset size.

\section{Conclusion}
\label{sec:conclusion}

We present visuomotor affordance learning (VAL), a method for learning tasks online in a new environment without supervision, utilizing trajectories from other related environments. VAL uses expressive generative models to learn visual affordances, combines these affordances with off-policy goal-conditioned RL to learn skills offline, and then fine-tunes in a new environment online. Like deep learning in domains such as computer vision~\cite{krizhevsky2012imagenet} and natural language processing~\cite{devlin2019bert} which has been driven by large datasets and generalization, robotics will likely require learning from a similar scale of data. In future work, VAL could enable such systems by allowing autonomous collection of coherent exploration data in diverse real-world settings.

\section{Acknowledgements}

This is an extended version of a paper by the same title that appears in the IEEE International Conference on Robotics and Automation (ICRA), 2021. This research was supported by the Office of Naval Research, the National Science Foundation through IIS-1651843, and Berkeley DeepDrive. We would like to additionally thank Misha Laskin, Vitchyr Pong, Glen Berseth, and Abhishek Gupta for constructive discussions about this work, and members of RAIL for their support and collaboration.

{ \small
\bibliographystyle{IEEEtranS}
\bibliography{example}
}

\clearpage
\newpage

\appendix

\subsection{Real-World Experimental Details} \label{sec:appendix_real}

\begin{figure}[b]
  \includegraphics[width=0.99\linewidth]{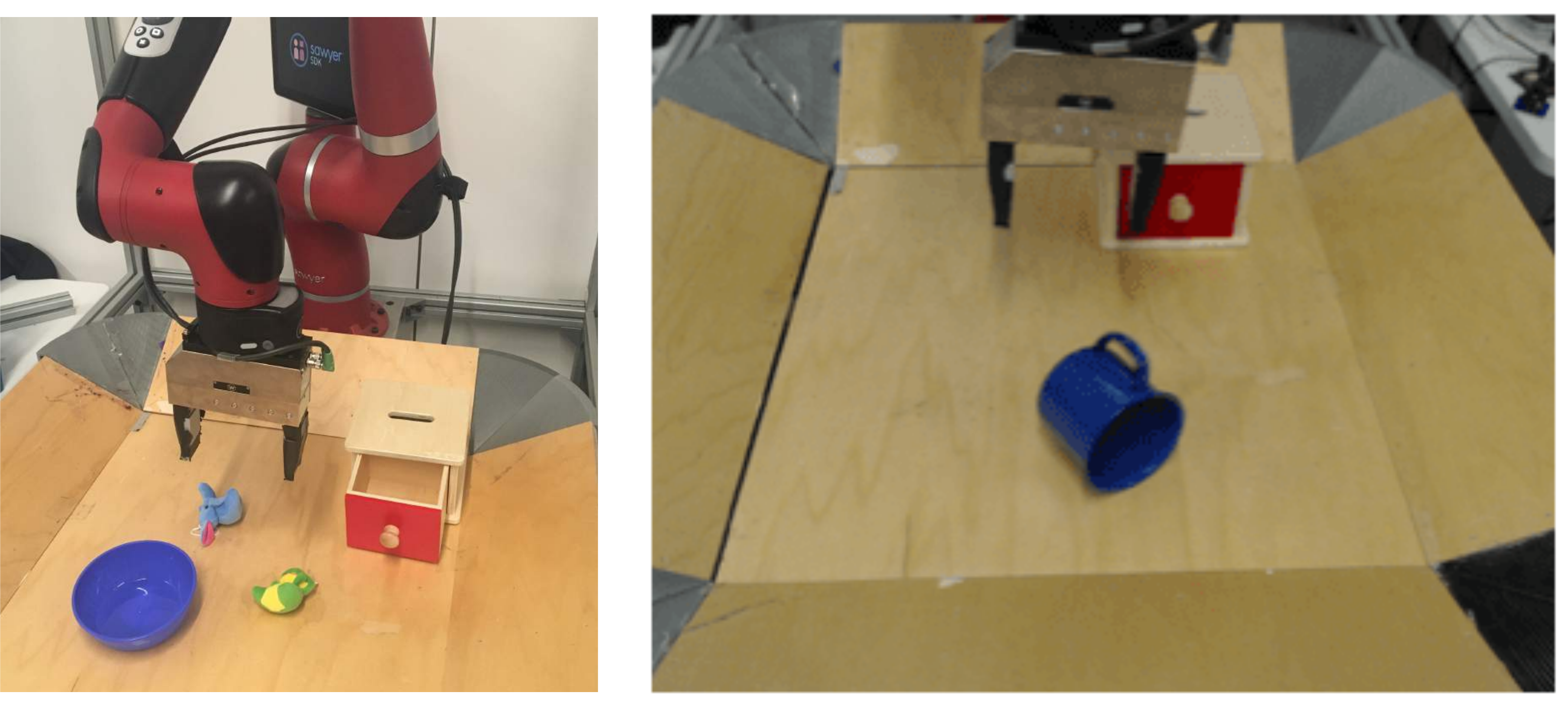}
  \caption{\small
  Left, full view of Sawyer robot setup. Right, camera view.
  }
  \label{fig:appendix_robot}
  \vspace{-0.5cm}
\end{figure}

Our real-world data used in experiments consists of 830 trajectories (61,482 transitions) collected by a human using a 3Dconnexion SpaceMouse device. Instructions for interfacing with the SpaceMouse is available publicly at \url{https://github.com/vitchyr/rlkit/tree/master/rlkit/demos/spacemouse}, with the device code adapted from the RoboSuite library~\cite{robosuite2020}. A full view of the robot and the view from the camera can be seen in Figure~\ref{fig:appendix_robot}.

Across our dataset, we interact with 10 drawer handles, 10 pot handles, 40 toys, and 60 distractor objects. Every 10 trajectories we randomly sample one or more interaction objects as well as two or more distractor objects. Before each rollout we randomize all object positions. The trajectories can be grouped into four separate categories: picking and placing toys, putting toys into a tray, opening a door and closing a drawer, and placing and removing a lid on a pot. To artificially increase the size of our dataset and make our policy robust to light changes and camera nudges, we utilize color jittering and random cropping during training. As there was an unequal amount of data per category, we re-balanced the dataset by using a different number of data augmentations per category. The final amount of task-specific data used per experiment is reported in Table~\ref{table:task-data-hyperparams}.

We additionally collected unscripted play data mixing all of the above behaviors with more object diversity, but did not use this data in this paper. With this data, there are 1,984 trajectories (137,111 transitions), covering 20 drawer handles, 20 pot handles, 60 toys, and 60 distractor objects. We also collected an additional 508 trajectories of on-policy robot data. The entire dataset is available on our website: \url{https://sites.google.com/view/val-rl}.

\begin{figure}
  \includegraphics[width=0.99\linewidth]{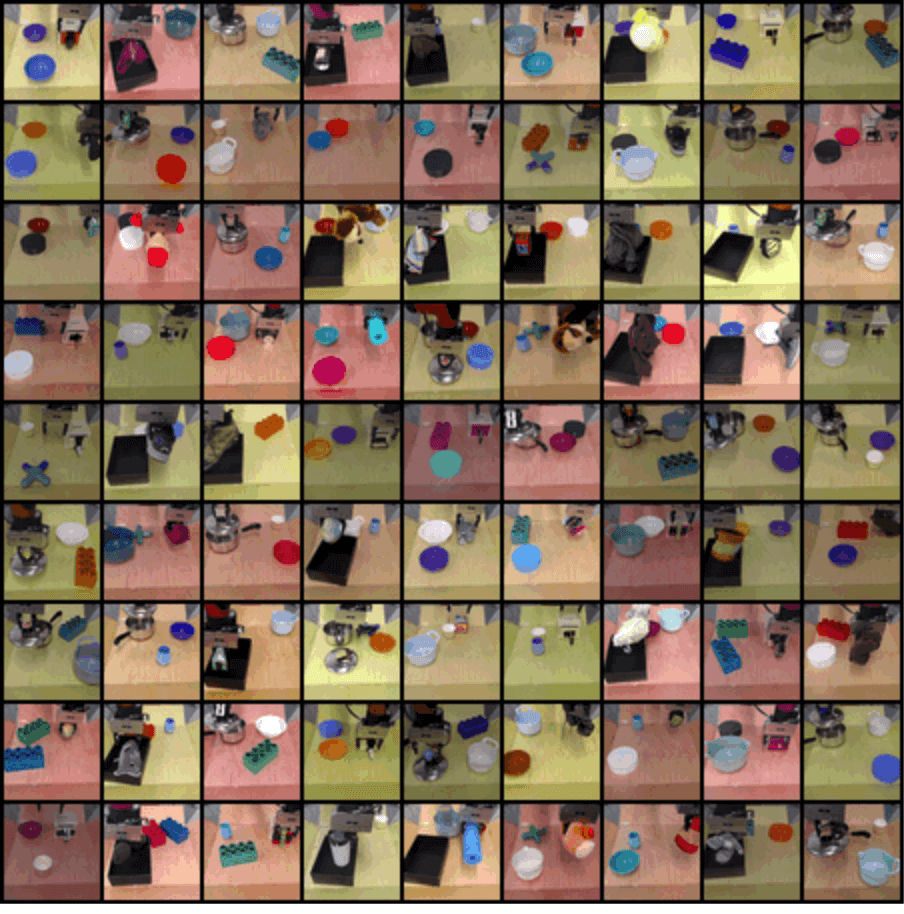}
  \caption{\small
  Images from the prior dataset. The dataset contains 830 trajectories. We have released the full prior dataset (containing 1,984 trajectories) along with on-policy robot executions at our website, \url{https://sites.google.com/view/val-rl}
  }
  \label{fig:appendix_dataset_imgs}
  \vspace{-0.5cm}
\end{figure}

\subsection{Simulation Experimental Details} \label{sec:appendix_sim}

Our simulated dataset consists of 8,000 trajectories (400,000 transitions). Before sampling each trajectory, we randomize the existence, position, color, and orientation of the following: two drawers, a box, a button, and an object. If an object is present it is chosen from a set of 84 object geometries. The trajectories are generated by a scripted policy which collects play data by interacting with all the present objects in a random order. The scripted behavior includes: opening and closing a drawer by the handle, opening and closing a different drawer by pressing a button, and re-positioning objects. All simulated RL experiments were run with 5 seeds.

\subsection{Algorithm Details}

Visuomotor affordance learning (VAL) builds off the \texttt{rlkit} codebase available at \url{https://github.com/vitchyr/rlkit}. We will release our code at our website, \url{https://sites.google.com/view/val-rl}. Below, we list the specific hyperparameters used in our experiments for each component. In VAL, we first collect an offline data $\mathcal{D}$, run representation learning, then offline RL, and finally online RL for a specific environment.

In the representation learning phase, we first train the VQVAE~\cite{oord2017vqvae} on $\mathcal{D}$. We then encode the entire dataset with the VQVAE to obtain discrete latent variables, and then independently train the PixelCNN~\cite{oord2016pixelcnn} on discrete latent code dataset. For the CCRIG experiments, we train a CCVAE~\cite{sohn2015cvae} on $\mathcal{D}$.

In the offline RL phase, we run advantage weighted actor critic (AWAC)~\cite{nair2020awac} on the offline data to obtain a single policy and Q-function. This policy and Q-function can then be fine-tuned to a specific environment by running online RL. 

All hyperparameters are provided below for these algorithms are provided below in tables~\ref{table:awac-hyperparams}, \ref{table:data-hyperparams}, \ref{table:vqvae-hyperparams}, \ref{table:pixelcnn-hyperparams}, \ref{table:ccvae-hyperparams}.

\begin{table}[h!]
    \centering
    \begin{tabular}{c|c}
    \hline
    \textbf{Hyper-parameter} & \textbf{Value} \\
    \hline
    Training Batches Per Timestep & $1$\\
    Exploration Noise & None (stochastic policy) \\
    RL Batch Size & $1024$ \\
    Discount Factor & $0.99$\\
    Reward Scaling & $1$\\
    Replay Buffer Size & $1000000$\\
    Number of pretraining steps & $25000$ \\
    Policy Hidden Sizes & $[256, 256, 256, 256]$\\
    Policy Hidden Activation & ReLU\\
    Policy Weight Decay & $10^{-4}$ \\
    Policy Learning Rate & $3 \times 10^{-4}$\\
    Q Hidden Sizes & $[256, 256]$\\
    Q Hidden Activation & ReLU\\
    Q Weight Decay & $0$ \\
    Q Learning Rate & $3 \times 10^{-4}$\\
    Target Network $\tau$ & $5\times10^{-3}$ \\
    Relabeling strategy $p_\text{RS}(\zt)$ & 50\% future, 30\% prior, 20\% rollout \\
    \hline
    \end{tabular}
\caption{Hyper-parameters used for RL (AWAC) experiments.}
\label{table:awac-hyperparams}
\end{table}

\begin{table}[h!]
    \centering
    \begin{tabular}{c|c}
    \hline
    \textbf{Hyper-parameter} & \textbf{Value} \\
    \hline
    Brightness (Color Jitter) & $[0.75,1.25]$\\
    Contrast (Color Jitter) & $[0.9,1.1]$\\
    Saturation (Color Jitter) & $[0.9,1.1]$\\
    Hue (Color Jitter) & $[-0.1,0.1]$\\
    Size (Random Resized Crop) & $48$\\
    Scale (Random Resized Crop) & $[0.9, 1.0]$\\
    Ratio (Random Resized Crop) & $[0.9, 1.1]$\\
    Interpolation (Random Resized Crop) & Antialiasing\\
    \hline
    \end{tabular}
\caption{Hyper-parameters used for data augmentation.}
\label{table:data-hyperparams}
\end{table}

\begin{table}[h!]
    \centering
    \begin{tabular}{c|c}
    \hline
    \textbf{Hyper-parameter} & \textbf{Value} \\
    \hline
    Convolution Layers & $3$\\
    Convolution Hidden Size & $128$\\
    Residual Layers & $3$\\
    Residual Hidden Size & $64$\\
    Embedding Size & $5$\\
    Dictionary Size & $512$\\
    Commitment Cost & $0.25$\\
    EMA Embedding & $False$\\
    \hline
    \end{tabular}
\caption{Hyper-parameters used for VQVAE training.}
\label{table:vqvae-hyperparams}
\end{table}

\begin{table}[h!]
    \centering
    \begin{tabular}{c|c}
    \hline
    \textbf{Hyper-parameter} & \textbf{Value} \\
    \hline
    Batch Size & $32$\\
    Layers & $15$\\
    Learning Rate & $0.0003$\\
    Latent Conditioning Type & $Continuous$\\
    \hline
    \end{tabular}
\caption{Hyper-parameters used for PixelCNN experiments.}
\label{table:pixelcnn-hyperparams}
\end{table}

\begin{table}[h!]
    \centering
    \begin{tabular}{c|c}
    \hline
    \textbf{Hyper-parameter} & \textbf{Value} \\
    \hline
    Convolution Layers & $3$\\
    Convolution Hidden Size & $128$\\
    Residual Layers & $3$\\
    Residual Hidden Size & $64$\\
    Embedding Size & $5$\\
    Conditioning Embedding Size & $1$\\
    \hline
    \end{tabular}
\caption{Hyper-parameters used for CCVAE training.}
\label{table:ccvae-hyperparams}
\end{table}

\begin{table}[h!]
    \centering
    \begin{tabular}{c|c}
    \hline
    \textbf{Hyper-parameter} & \textbf{Value} \\
    \hline
    Tray & $25 \%$ task specific data\\
    Pick and Place & $25 \%$ task specific data\\
    Close Drawer & $20 \%$ task specific data\\
    Open Drawer & $20 \%$ task specific data\\
    Place Lid & $17 \%$ task specific data\\
    \hline
    \end{tabular}
\caption{Hyper-parameters used for task-specific replay buffer re-balancing (through data augmentation).}
\label{table:task-data-hyperparams}
\end{table}

\end{document}